# Categorizing ancient documents

Nizar ZAGHDEN[1],
Remy MULLOT[2], Mohamed Adel ALIMI[3]

1, 3  REGIM: Research Group on Intelligent Machines,
University of Sfax, ENIS, Department of Electrical Engineering.
BP W - 3038, Sfax, Tunisia
*nizar.zaghden@gmail.com, adel.alimi @ieee.org*
2   L3I: Laboratoire Informatique Image Interaction,
Université de La Rochelle, France BP 17042
*remy.mullot@univ-lr.fr*

**Abstract**
The analysis of historical documents is still a topical issue given the importance of information that can be extracted and also the importance given by the institutions to preserve their heritage.
The main idea in order to characterize the content of the images of ancient documents after attempting to clean the image is segmented blocks texts from the same image and tries to find similar blocks in either the same image or the entire image database. Most approaches of offline handwriting recognition proceed by segmenting words into smaller pieces (usually characters) which are recognized separately. Recognition of a word then requires the recognition of all characters (OCR) that compose it. Our work focuses mainly on the characterization of classes in images of old documents. We use Som toolbox for finding classes in documents. We applied also fractal dimensions and points of interest to categorize and match ancient documents.

## 1. Introduction

The analysis of historical documents is still a topical issue given the importance of information that can be extracted and also the importance given by the institutions to preserve their heritage. Typically an analysis of ancient documents is preceded by a cleaning phase of noisy images. The preprocessing or enhancement of the quality of the images of ancient documents seems important. This subject is vast and there are a variety of methods and techniques that have been applied in order to achieve analysis or extracting information from images of old documents. Characterization of content or indexing [1] and text search works from ancient documents [2] are examples of applications that we can apply on the old documents. In fact before choosing to apply such an application on ancient documents we must first target the audience who will use our application. Such applications can be divided into three categories : Librarians who want to find printed ancient history, students or researchers. In this paper, we are interested in categorizing the content of images of old documents. We apply several methods in order to characterize the content of images of old documents. Madonna bases, library Tunisian national [3], the British library and also manuscripts of George Washington will be the scope of our application (http://www.wdl.org/fr/).We do not believe this until we can develop a system capable of characterizing efficiently and equally different images from our heterogeneous database, however we try to find systems that are more adaptable to each of the bases on which we work. Similarly, we believe that the techniques based on the segmentation of text character remain valid for printed documents (Madonna), but are not directly used for handwritten bases (manuscripts of George Washington) or for the Arabic script is cursive in nature. To characterize the classes contained in the same images we made use K-means classifier with which we were able to identify the different classes belonging to our image shown in Figure 1(fig 1). K-means in effect is a tool Classical classification that divides a set of data into classes homogeneous. Most images locally verify properties of homogeneity, particularly in terms of light intensity. The following figure shows in fact the application of k-means on a document image old. Five different figures represent the five different classes extracted by this algorithm. The major drawback of k-means is the number of classes is chosen in advance.

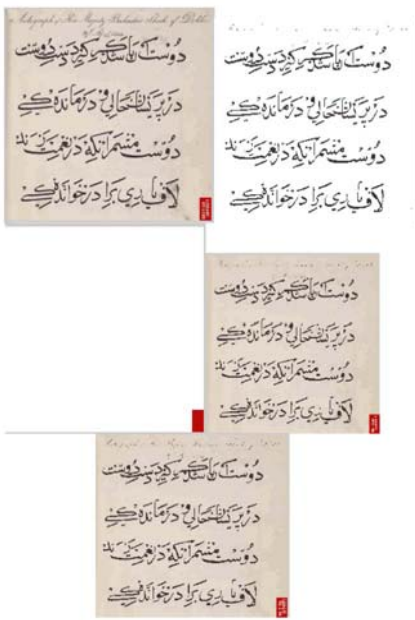

Fig. 1 Different classes of an image after application of k-means

This can lead, in fact, or the overlapping of two classes (number of classes priori selected is less than the number of classes in the image), or the excess of classes that can be classified in an image. This exceeded classes may be for example in an image that has only two classes while we selected a number of classes in advance of four or five which is well above the classes may be contained in the image (this is a bit similar to the phenomenon of "over-segmentation "estimated by Al Badr in [4]).We also used the toolbox "SOM" [5] to categorize the documents. The results we found are relatively average (Fig 1) and this is mainly because we work with document images which we cannot estimate the number of classes that can be contained in the same page. And thus to remedy the defect of these methods, it would be more reasonable to work with unsupervised classification tools because we cannot estimate in advance the number classes in any possible basis on which we try to categorize. Indeed the k-means algorithm is used to determine the various classes and from an image these images we can find a picture in the foreground, background image.

## 2. Characterization of ancient documents by fractal dimensions

Several methods have been developed to calculate the fractal dimension as method of counting box (Box-Counting) [6] and the method of dilation. The Box-Counting dimension is considered the simplest method for calculating the fractal dimension and is applied to pictures that are not empty.

From the properties of the self-similarity, fractal dimension of a set A is defined in Eq. (1):

$$D = \log(N) / \log(1/r) \qquad (1)$$

N is the total number of separate copies similar to A, and $1/r$ is the scale factor with which A is divided. We opted to use several methods to characterize fonts. First we have chosen to apply the fractal dimension, major feature of fractal geometry. For this we implemented two algorithms on the method of counting box and an algorithm for the method of dilation.

In fact the method of counting box (Box-Counting) is based on the size of Hausdorff. As for the method of dilation, it is based on the dimension of Minkowski

Bouligand [7]. The latter is in fact an increase of the Hausdorff dimension [7].

The results obtained with the fractal dimension show the discriminating power of this method in recognizing printed Arabic fonts. In fact, the calculation of fractal dimensions of images is simple and requires no preprocessing step of images of writing. We also compared the algorithms of fractal dimensions that we have developed with other methods on the same images and textures. We noticed the clear superiority of the new approach that we have adopted (Counting Density per Box: CDB).

The performance of this method in noised images pushed us to test this method in old documents. Indeed, the fractal dimensions from the CDB method for resolutions greater than 60 dpi achieved rates recognition almost similar to the results obtained with the same approach for clear images. The results for the variation of resolution are shown in table 1.

Table 1: Font recognition rate for various resolutions

| *Resolution* | 300 | 200 | 100 | 75 | 60 | 50 | 40 |
|---|---|---|---|---|---|---|---|
| *Rate (%)* | 96.5 | 96.5 | 96.5 | 96 | 95 | 85.5 | 80 |

In terms of recognition rates, we noticed a significant performance the new method we have adopted for the calculation of the fractal dimension.

In fact, this method alone has achieved a recognition rate of 96.5%. The results show that the methods we have adopted for the recognition of multifont texts can be applied to develop a robust system of character recognition.

Indeed the methods we have developed can be applied to a number of larger fonts and from heterogeneous databases with a variation of resolution and degradation non uniform noise as if images of old documents.

In the state of the art collection of images with tiles, there is different practices recovery, and various studies show that they lead to results.

The determination method called "method of boxes", commonly used in analysis of images, using a grid of the map mesh much finer (fig 2).

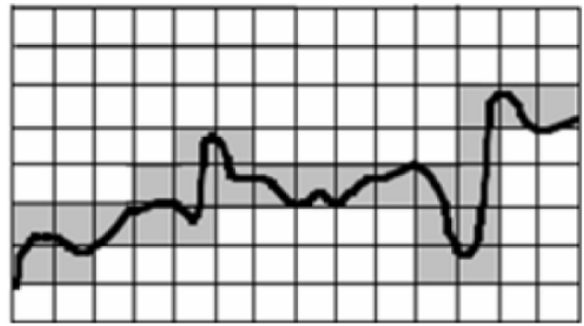

a: recovering object with square of size l

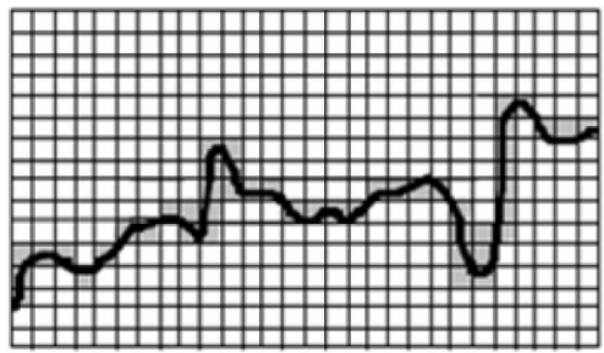

b: recovering object with square of size l/2

Fig. 2 Box counting method

The new method we developed to calculate the fractal dimension is inspired by the box counting method [8]. Among the other derivatives this method, there is the method of counting reticular cells as the Box-Counting differential [6].

## 3. Characterization of ancient documents by local approaches

Characterization and interpretation of the content of textured images is a problem important for many application areas. Many works based on statistical descriptors but recent work has demonstrated the relevance of the approaches based on the extraction and characterization of points of interest [9]. We studied two types of local approaches: Points of Interest methods and pixel by pixel methods.

The interest of points of interest approaches lies in their properties including invariance change of contrast and affine images. A region of an image comprises characteristic points of an image carrier in particular information, called points of interest. Such points belong outline of a picture: we speak in this case of contour points.

To overcome the problems of characterization of pages of documents (execution time, computation time of the feature vectors ...), it was necessary to move towards a new methodology to accurately locate the text areas, identify shapes redundant without using any segmentation of the page and thus leading to a form intelligent partition of the image centered around the interesting areas. Figure 3 shows the points of interest of a line of text (fig. a) as well as points of interest image (Fig. b).

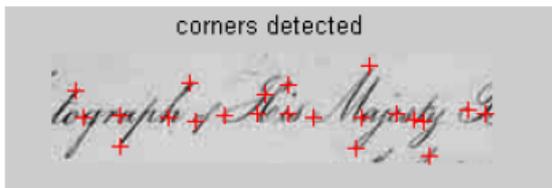

a: points of interests for a text line

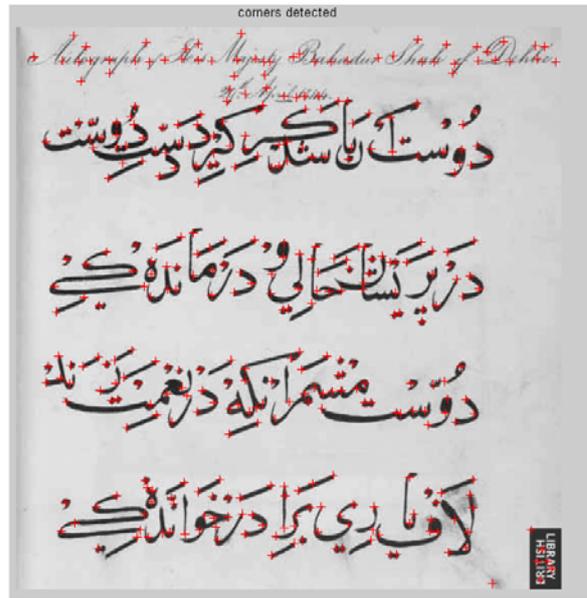

b: image points of interest

Fig. 3 Detection of points of interest on images of ancient documents by the SIFT method

We can recall here that the partitioning of the image is the basis of many methods and recognition that can be achieved ideally through research guided by ROIs.

In the context of global characterization of words, an approach to the description of local points of interest in an image was studied. It allows information to calculate a local level of the contour points of a region of interest of the image. All these local information can then calculate a signature representative of the region of interest of the image, to identify the form in its entirety. In this context, it is to measure the distances between shapes by matching points between two images, that is to say, the recognition of the points in common between at least two digital images. We compare two points of interest images based on "British Library" function from the "XOR" (Fig 4).

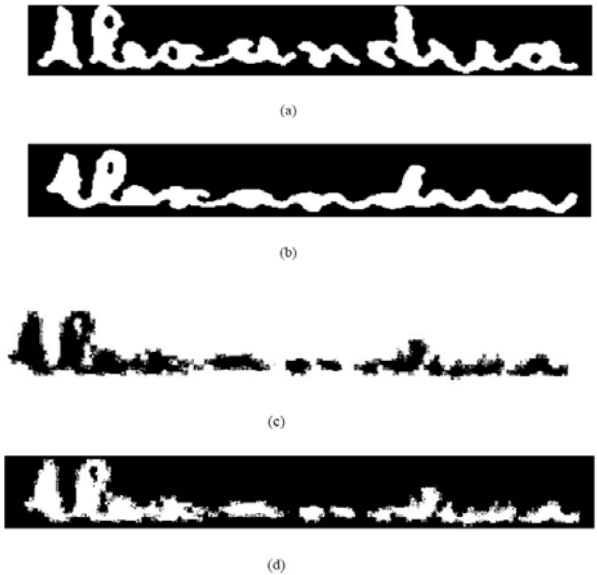

Fig. 4 Applying XOR for two occurrences of the word "Alexandria"

Thus, we can characterize different text blocks of the same size with the function XOR, in which we represent in the resulting image a white pixel if it is a white pixel in one of the first two images, and it is the opposite in the other file. So XOR do the comparison pixel by pixel. Calculating the similarity between two shapes occurs from the symmetric differences between the new form and the model images (c) and (d) of the previous figure are the result of the XOR function on both Photo (a) and (b). In Figure 5 we present the result of the XOR function of two occurrences of the same line( fig 5). Thus we can conclude that the XOR function can we solve the problem of word recognition images of old documents when forms are strictly identical and repositioned relative to each other.

In fact, we have applied the Euclidean distance map (EDM algorithm: Euclidian distance map) to calculate the error between images. In fact, each white pixel in the resulting image corresponds to an error and the EDM function applied to XOR allows us to obtain a vector measuring the error between two images. This basic approach has many disadvantages that make it use very difficult because the characterization by points of interest can bring sometimes non similar words even if they have similar characteristics.

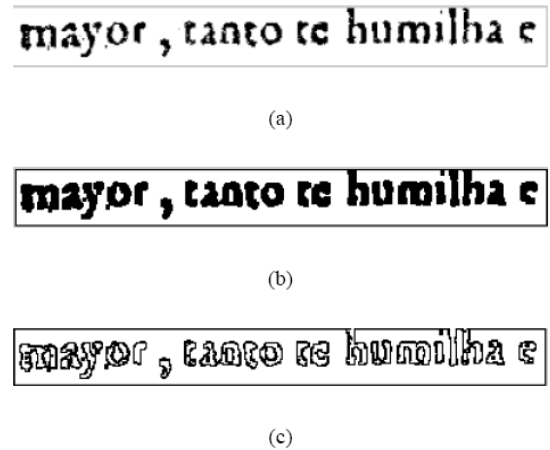

Fig. 5 Applying XOR for two occurrences of the same line

3.1 Application on pseudowords: wordspotting

Pixel by pixel methods is also studied here illustrated by the wordspotting method, which can actually perform the matching of words pictures of old documents. The only problem resided in fact on the basis of images that we have. George Washington's manuscripts have many redundant words and many researchers have worked on it. The contribution that we wanted to do is not to work with a collection of words, but to work with images characters or blocs of characters.

Documents on which we conduct our approach are only old documents prints. Indeed, we perform an initial image segmentation based on the method developed in [10] to group blocks of text pictures. Then we treat each block choosing as a criterion for separation between the characters to be segmented, the presence of white pixels. Admittedly, this approach allows the correspondence between characters on old documents printed but will also characterize pseudowords (blobs) from ancient manuscripts including Arabic characters printed or manuscripts because of the nature of the Arabic script, which is cursive in nature [11]. In the first place we chose to work with the XOR function followed by the algorithm EDM to calculate the errors between the query images and images of the base (Fig 6). Images (c) and (d) respectively and the results complement the results of the XOR function applied to the frames (a) and (b) (Figure 6).

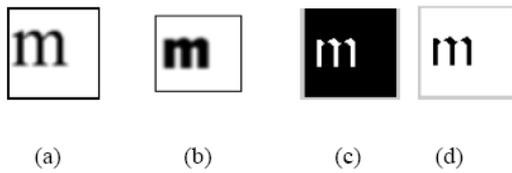

Fig 6. Applying the XOR function of the two images character "m"

Secondly, we calculated the projection profile character images (Figure 7).The result of each character is actually compared with the other characters in the document. One of the main advantages is that we do not know the number of classes or characters in advance whether in a document image or any former base.

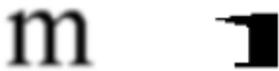

Fig 7. Vertical profile of the character "m".

3.2 Performance of wordspotting

To evaluate different algorithms for image retrieval by content, we require a measure of performance. There are several types of performance measurement such as PWH "Weighted Percentage of Hits," RPP "Recall and Precision Pair" and PSR "Percentage of Similarity Ranking". We will detail the RPP, which defines the measures of precision and recall. Performance research is evaluated using precision and recall. Precision P is defined as is the ratio of the number of relevant images r to the desired total number of images n found it therefore measures the system's ability to find relevant images. The point R is defined as the number of images on the relevant desired number m total relevant images in the entire database, so it measures the ability of system to find only relevant images. Precision P and recall R are given in Eq. (2):

$P = r / n$ and $r = R / m$ (2)

To measure these parameters, we assume that the image database is composed of classes disjointed images where we know the Cardinals respective "m". All pictures the base are taken successively as images and queries we see images returned to a rank n variable. For setting a reminder, we draw curves "

$R = f(n)$. " Faster the curve tends to 1 and the system will perform.

For parameter precision, we draw curves "$P = f(n)$." The system is more accurate and more P curve decreases slowly.

The images that we have applied our algorithms are those from the base Madonna and gutenberg since the basis of the national library contains much more noise and that the techniques we developed are not sufficiently effective on such images also degraded. Other studies related to the processing of old documents interested firstly on the identification phase defects on ancient documents. We chosen in our case study work with the Gaussian filter in the above objective eliminate pixels away, but throughout our report we avoid making the preprocessing of document images for causes: on the one hand with some pretreatments, we will lose the information of images of old documents that may be relevant for the description and characterization of this image. On the other hand the noise varies from a document image old compared to another, so that we do not let the choice of finding suitable filters to remove noise generated in a base heterogeneous, unless it is the same basic container may be the same defects (Tasks ink and water stains or other).

Our experiments have touched only the images of Latin characters whose number is 7280. Then we tested images of words in printed documents which number is 100. These test images are repeated 135 times in the images seen in our learning base (there are words that are repeated more than once). We have calculated the terms "precision" and "recall" for the two different methods we have adopted:

- XOR
- EDM (after applying XOR)
- Vertical projection followed by EDM algorithm

We present in Table 2 the results obtained with the three methods described.

Table 2: Precision rate and recall methods adopted

| *Method* | XOR | EDM | Projection profile |
|---|---|---|---|
| *Precision (%)* | 63.14 | 78.43 | 41.67 |
| *Recall (%)* | 55.34 | 79.32 | 34.46 |

We can also notice that the best recognition rates were obtained with EDM method. While this rate is still high in comparison with the methods wehave tested but it will certainly be improved by adding other criteria.

Among the works that we are making now, there are correct and improving the inclination of characters from old documents. Other techniques are also our present field of study such as DTW (Dynamic Time Wrapping) function used in the analysis of time signals, but we will use in our field study.

Despite their effectiveness, local methods have the following disadvantages:
• calculation time prohibitive depending on the size of the analysis window;
• over-segmentation faults and paper texture on the background of the image;
• treatment of difficult documents font size is very variable,
• analysis window is fixed throughout the treatment.

To address the problem of variability in the size of the images to be compared, we use the detector points of interest. We tried mainly the Harris detector [Har 98] and the SIFT descriptor [13]. The first detector can bring good results especially for images standardized; however it fails for images of varying size. The solution is to use detectors points of interest such as multi-scale SIFT, which is the most robust sensor currently in the literature [14].

3.3 Application of points of interest on the entire of the document image

Generally, a matching algorithm based on the points of interest is divided into three parts:

1) detection of points of interest (Harris, difference of Gaussians in SIFT ...)
2) characterization of points of interest (eg each point of interest, we associate a vector: pixel values in a neighborhood, local jet (successive derivatives), ...) or even a histogram (SIFT).
3) Mapping. To simplify, just from a distance (Euclidean Mahanalobis, ...) and measure the distances between the vectors of points of interest from the image 1 and those of the image 2. If the distance is smaller than a certain threshold we set, then we put mapping the points.

Harris detector [12] is a corner detector and it is not multi-scale. That is to say that if we compare two images, one of which is a zoom of the other using the Harris detector, the rate of correspondence will be weakened due to the property of detector "non multi-scale."

The solution is to adopt a multi-scale detector such as SIFT, which has given best results compared to other detectors [14]. Characterization is to find information on a point of interest other than its position. In what follows, we will detail the steps taken to make the correspondence pictures of old documents based on the sensor point of interest: SIFT [13].

As a first step we work with hundreds of pictures of old documents to see the results and then make the appropriate steps according to the results if they are interesting or not.

As the number of images (100 images) on which we performed correspondence is high, we applied a first phase of rejection.

This phase is based on characterizations overall image by the application of the fractal dimension. Indeed, we have shown previously that this technique can achieve distinction or grouping of different classes of images, but it remains still insufficient to match the images of old documents efficiently. This is why we chose to combine those techniques to local overall to get the best results from characterization of the images of old documents. We illustrate in Fig 8 the overall scheme of this system [15].

We applied our system to approximately 1000 images from our database because treatments preprocessing and normalization. We calculated fractal dimensions for all images from our database. The choice of method is fractal due to its discriminating power but also to the gain in computation time compared with other global methods [16]. In the image below, we present the principle of elimination a set of images with fractal dimensions above a certain threshold manually at first.

In the second step, we consider just the images belonging to group 1 and group 2. We use these images to the SIFT descriptor for target images that best match the query image. This is very useful because we working with a large number of images. The number of resulting images is not always fixed, but depends mainly on the query image and also the number of images similar to the query image.

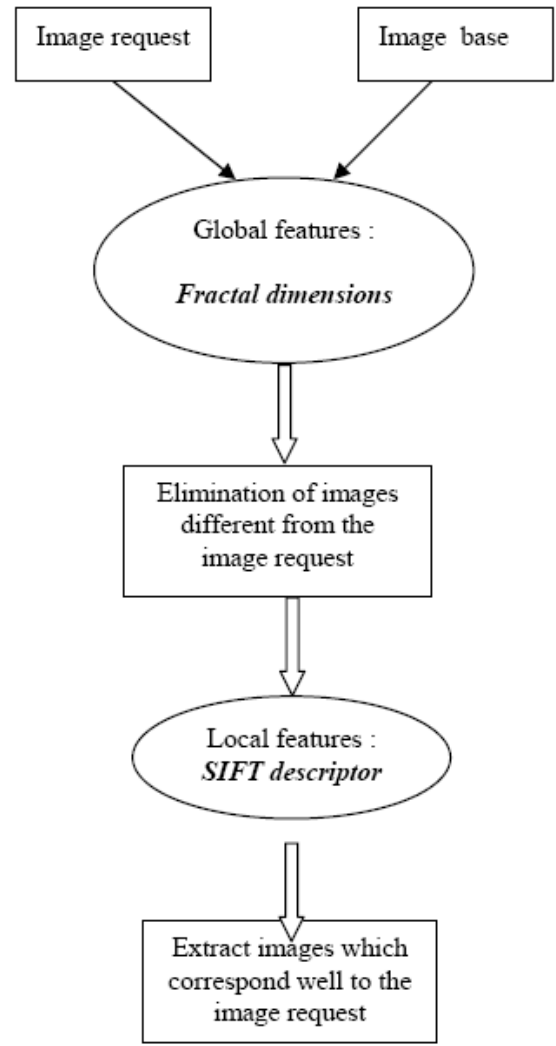

Fig 8. Global scheme of the method

In the second step, we consider just the images belonging to group 1 and group 2. We use these images to the SIFT descriptor for target images that best match the query image. This is very useful because we working with a large number of images even if they are carried out from Arabic and Latin bases [17]. The number of resulting images is not always fixed, but depends mainly on the query image and also the number of images similar to the query image.

We performed a preprocessing phase on 1000 images from our database in order to work with normalized images with a resolution of 300 dpi and the same size: 512

* 512 pixels. The system adopted here give good results compared with other methods in time of execution and number of iterations, but it can be more precise if we apply other parameters issued from fuzzy logic [18] to add the membership degree of each class.

## 4. Conclusion

We applied several experiments to categorize the content of images old documents. The determination of number of classes in an image document was achieved by K-means. We tried also to categorize images through a global approach with the fractal dimension. We were able to identify portions of images belonging to the same classes. Given the variability of the content of the images we are dealing with, as well as their huge number, we decided to combine the approaches with holistic approaches local descriptors using points of interest.

With the Harris detector, we could apply the wordspotting on the pseudowords images of ancient documents, then we calculated the errors between the query images and target images using the XOR function and algorithm "Euclidian Distance Map» (EDM). The disadvantage of this method is mainly the execution time, since we work with local approaches on a huge image database. We also applied the SIFT descriptor with images of old documents. This phase was carried out after the first stage of rejection distant images of the query image based on the fractal dimensions. The choice of this approach is due to its overall discriminating power, but also to the execution time which is rapid in comparison with the other methods.

**Nizar ZAGHDEN** was born in sfax (1978). He received the B.S. degree in computer science from The sciences faculty (FSS) in 2003, and the Master degrees in Computer Science from the National Engineering School of Sfax - Tunisia (ENIS), in 2005. On march 2013, he became a philosiphate Doctorate after defending thesis. He became a doctorate in computer science. He joined the Sfax University (USS), where he was an assistant professor in the Department of Computer science of the management faculty of sfax (Gestion). He is now attached to the university of Gabes where he works as assistant in the higher institute of computer science in Medenine.

He is member of the REsearch Group on Intelligent Machines (REGIM). His research interests include Computer Vision and Image and video analysis. These research activities are centered around document analysis and Pattern Recognition. He is a Graduate Student member of IEEE. He was the member of the organization committee of the International Conference on Machine Intelligence ACIDCA-ICMI'2005.

**Rémy MULLOT** is a Professor and Director of the L3I laboratory in La Rochelle in France. His research activities are placed in the context of recognition and indexing documents, since these processes are associated with a content analysis of the image. The targets are very wide because they can focus on ancient documents for which the laboratory has developed real expertise, but more broadly to any type of highly structured materials (mainly based on textual components) or low structured (mainly based on graphical data). These research themes are very cross because they are both extractors indices (signatures) on treatment systems, through the categorization of content, structures, styles.

These activities are opportunities direct projects larger laboratory in conjunction with other skills in L3i: ontologies and images.

**Adel M. Alimi** was born in Sfax (Tunisia) in 1966. He graduated in Electrical Engineering 1990, obtained a PhD and then an HDR both in Electrical & Computer Engineering in 1995 and 2000 respectively. He is now professor in Electrical & Computer Engineering at the University of Sfax.

His research interest includes applications of intelligent methods (neural networks, fuzzy logic, evolutionary algorithms) to pattern recognition, robotic systems, vision systems, and industrial processes. He focuses his research on intelligent pattern recognition, learning, analysis and intelligent control of large scale complex systems.

He is associate editor and member of the editorial board of many international scientific journals (e.g. "Pattern Recognition Letters", "NeuroComputing", "Neural Processing Letters", "International Journal of Image and Graphics", "Neural Computing and Applications", "International Journal of Robotics and Automation", "International Journal of Systems Science", etc.).

He was guest editor of several special issues of international journals (e.g. Fuzzy Sets & Systems, Soft Computing, Journal of Decision Systems, Integrated Computer Aided Engineering, Systems Analysis Modelling and Simulations).

He was the general chairman of the International Conference on Machine Intelligence ACIDCA-ICMI'2005 & 2000.

He is an IEEE senior member and member of IAPR, INNS and PRS. He is the 2009-2010 IEEE Tunisia Section Treasurer, the 2009-2010 IEEE Computational Intelligence Society Tunisia Chapter Chair, the 2011 IEEE Sfax Subsection, the 2010-2011 IEEE Computer Society Tunisia Chair, the 2011 IEEE Systems, Man, and Cybernetics Tunisia Chapter, the SMCS corresponding member of the IEEE Committee on Earth Observation, and the IEEE Counselor of the ENIS Student Branch.